\documentclass[11pt]{article}

\usepackage[preprint]{acl}

\usepackage{times}
\usepackage{latexsym}
\usepackage{tabularx}
\usepackage{float}
\usepackage{placeins}
\usepackage[T1]{fontenc}

\usepackage[utf8]{inputenc}

\usepackage{makecell}
\usepackage{rotating}
\usepackage{tikz}

\usepackage{microtype}

\usepackage{inconsolata}

\usepackage{graphicx}
 \usepackage{booktabs}
\usepackage{amsmath}
\usepackage{amssymb}   
\usepackage{url}
\usepackage{hyperref}

\makeatletter
\ifacl@anonymize
  \def\githuburl{https://anonymous.4open.science/r/hal-0B35}
  \def\uniname{[Anonymous US University]}
\else
  \def\githuburl{https://github.com/ROC-HCI/hal}
  \def\uniname{University of Rochester}
\fi
\makeatother

\title{HAL: Inducing Human-likeness in LLMs with Alignment}
\author{
Masum Hasan \quad
Junjie Zhao \quad
Ehsan Hoque \\
University of Rochester\\
\texttt{\{m.hasan@, jzhao58@u., mehoque@cs.\}rochester.edu}
}

\begin{document}
\maketitle

\begin{abstract}

Aligning language models to qualitative behavioral traits, such as human-likeness, remains difficult because they are hard to define, measure, and optimize. As a result, improvements in human-like behavior are largely driven by scale or broad supervised training, rather than targeted alignment. We introduce Human Aligning LLMs (HAL), a framework for aligning language models to conversational human-likeness using an interpretable, data-driven reward. HAL derives explicit conversational traits from contrastive dialogue data, combines them into a compact scalar score, and uses this score as a transparent reward signal for alignment with standard preference optimization methods. Using this approach, we align models of varying sizes without affecting their overall performance. In large-scale Chatbot Arena-style human evaluations, a model aligned with HAL is more frequently perceived as human-like in conversation. Because HAL operates over explicit, interpretable traits, it enables inspection of alignment behavior and diagnosis of unintended effects. More broadly, HAL demonstrates how soft, qualitative properties of language--previously outside the scope for alignment--can be made measurable and aligned in an interpretable and explainable way.
\end{abstract}

\section{Introduction}

\begin{figure*}
    \centering
    \includegraphics[width=\linewidth]{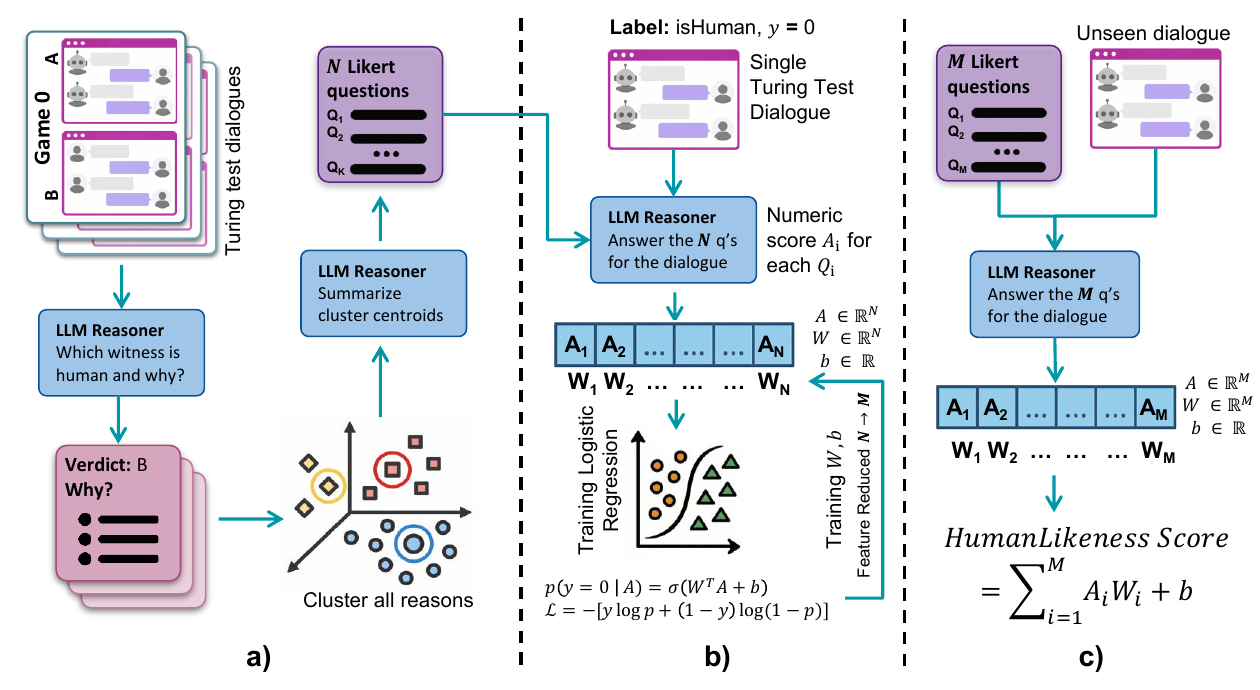}
    \caption{The HAL pipeline: (a) identifying human-likeness traits from contrastive dialogues (e.g. Turing tests), (b) learning trait weights via a proxy classification task, and (c) computing a human-likeness score for alignment.}
    \label{fig:main_figure}
\end{figure*}


 Human communication is the product of millions of years of social evolution, shaped by subtle and largely unspoken norms. While these norms are difficult to articulate; however, once broken, easily detected. When artificial agents fail to reproduce them, interactions can feel mechanical or uncanny.

Human-like conversational behavior is especially important in settings where interaction quality matters more than task completion. These include role-play and character simulation \cite{roleplay1, roleplay2, roleplay3, roleplay4}, communication training \cite{comm1, comm2, sapien}, patient simulation in healthcare and mental health contexts \cite{health1, sophie1.0, health2, ptsd, health3}, and others.

Despite its importance, conversational human-likeness remains difficult to define and even harder to measure. Humans can often tell whether a conversational partner is human or artificial, but this judgment is typically holistic and implicit rather than based on explicit criteria. As a result, there has been no systematic way to measure human-likeness, and consequently no clear reward signal for aligning models toward it. This has left training language models to be more human-like largely out of reach for alignment research.

Recent large-scale Turing test results highlight both progress and limitations \cite{llm-turing}. \texttt{GPT-4.5} was judged to be human in 73\% of comparisons, while \texttt{LLaMA-3.1-405B} achieved near-chance performance and smaller baselines performed far worse. These findings suggest that scale can improve perceived human-likeness, but they do not explain why, nor do they provide a clear recipe for training models to be more human-like.

In this paper, we introduce \emph{Human Aligning LLMs} (HAL), a framework for quantifying conversational human-likeness and using it as a reward for alignment. Our approach is entirely data-driven: we extract recurring human-likeness cues from contrastive dialogue data (e.g., Turing tests), compress them into a compact and interpretable set of traits, and combine them into a single scalar score. We then use this score as a reward signal for alignment with standard preference optimization methods such as Direct Preference Optimization (DPO) \cite{dpo}. Across models of varying sizes, we show that alignment with HAL leads to clear improvements in perceived human-likeness under human evaluation, while largely preserving performance on other benchmarks.

Concretely, our proposed framework HAL:
\begin{enumerate}
    \item identifies recurring conversational traits that reliably distinguish human--human from human--AI dialogue,
    \item compresses these traits into an interpretable, scalar measure of conversational human-likeness,
    \item uses this measure as a reward signal for alignment with standard preference optimization methods, and
    \item demonstrates through human evaluation that models aligned with HAL are more frequently perceived as human-like.
\end{enumerate}

More broadly, HAL offers a general methodology for inducing soft, qualitative traits in language models—traits that are difficult to specify directly, but can be inferred from contrastive data. By making such traits measurable and interpretable, this work opens new possibilities for controllable, transparent, and human-centered alignment beyond conventional objectives.

Our best weights are available on HuggingFace and Ollama. Our training code and datasets are available at: \url{\githuburl}
\section{What Makes Human Conversation Human?}
\label{sec:properties}

We begin with a simple premise: to make a model more ``human-like'' in conversation, we should first understand which conversational cues best help identify it. Once we are able to understand and quantify that, we can align the model to demonstrate more of that behavior.

\subsection{Characteristics of Human-likeness From Turing Tests}
\label{subsec:turing}

The Turing test is an imperfect proxy for human-likeness, but it provides two ingredients that are difficult to obtain otherwise: paired dialogues that are designed to be compared, and which side is human. We analyze the Turing test dialogue transcripts released by \citet{llm-turing}. 

The original dataset consists of 1116 games, where the \emph{investigator} $I$ interacts with two unknown conversational partners (the \emph{witnesses} $W$'s), and then decides which $W$ is human. Each game consists of two conversations between $I$ and two $W$'s, along with (i) the ground truth, (ii) the investigator's decision, and (iii) a brief free-form explanation of the decision. We filter out games where either conversation is $<50$ words, as short conversations are either obvious or uninformative. Hence, we are left with 557 games, each consisting of a human-AI and a human-human conversation. In our filtered dataset, the human judge has an accuracy of $54.58\%$ in identifying correctly who is human, which is slightly better than random.

\paragraph{LLMs as Turing judge.}
A natural starting point is to treat the investigators' free-form explanations as a source of human-likeness cues. In practice, they are inconsistent in format and very frequently incorrect.
We therefore construct an \emph{LLM-based Turing judge} that evaluates the same paired dialogues, but produces explicit and structured reasons. For each Turing test game, we pass the judge with two dialogues in random order, ask it to (i) predict which witness is human, and (ii) provide 3-5 Likert-style statements that helped make the decision on this specific pair of dialogues (full prompt at Appendix \ref{fig:prompt-find-statement}).

The goal of this experiment is not to claim a new best Turing judge, but to identify a set of high-signal descriptions of human-likeness cues using the classification accuracy as a proxy for the quality of the reasons. We evaluate a cohort of commercial and open-source models as displayed in Table \ref{tab:turing-acc}. In our experiments, \texttt{GPT-5} with \texttt{high} reasoning attained the highest accuracy of $64.81\%$, and therefore its ``reasons" are used for further analysis.

\paragraph{Creating a compact set of characteristics.}
Running the LLM-judge over the dataset yields $2,735$ total natural-language reason statements, many of which are redundant. We embed each reason statement using a sentence encoder \cite{wang2020minilm} and cluster them using a density-based clustering algorithm \cite{HDBSCAN}. We find 53 representative clusters and extract their centroid. The 53 centroid statements still contained redundancies. Hence, we further instruct an LLM to summarize the 53 clusters into distinct Likert-style statements (prompt Appendix Figure \ref{fig:prompt-cluster-summary}). This yields a final inventory of 32 characteristics, presented in Appendix Table~\ref{tab:hl32q}. Henceforth, we refer to these 32 traits as \emph{Human-Like 32 Questions} or HL32Q.

The resulting characteristics reflect what repeatedly distinguishes human and model dialogues in this dataset under a Turing-style comparison. This pipeline is visualized in Figure \ref{fig:main_figure} a).

\section{Quantifying Human-likeness}
\label{sec:quantifying}

In previous section, we identified a compact set of conversational traits (HL32Q) that repeatedly distinguish human--human from human--AI dialogue in a Turing-style setting. Our next goal is to turn these qualitative traits into a quantitative signal and derive a single score to a dialogue that reflects how human-like it appears. This simple score will be used as a reward signal in alignment training. To extract quantitative signals from qualitative statements, we draw on survey methods developed in social science and psychology.

\subsection{Human-likeness Classifier}
\label{subsec:classifier}

Given a dialogue, an LLM judge (\emph{HL32Q Judge}) rates its agreement with each of the 32 statements only based on the witness responses on a 1--5 Likert scale (full prompt Appendix Figure \ref{fig:prompt-hl-judge}). Unlike the Turing test setting, this is not a pairwise comparison. Each dialogue is scored independently, producing a fixed-length feature vector $\mathbf{A} \in \mathbb{R}^{32}$. This representation compresses a dialogue into a small number of high-level and complex conversational cues.

We propose a simple method to compress this rich vector $\mathbf{A}$ into a single scalar human-likeness score. We train a logistic regression model that takes $\mathbf{A}$ as input and predicts $y$, indicating whether the witness in the original dialogue is human ($y=1$) or an AI ($y=0$). By training with backpropagation, we learn the weight matrix $\mathbf{W}$ and add bias $\mathbf{b}$. The model learns a linear decision boundary over the $32$ features, yielding a weight for each trait that reflects its contribution to distinguishing human from AI dialogue. Even though more accurate aggregation methods can be developed, we focus on simplicity and interpretability in the current study.



Formally,
\begin{equation}
\begin{gathered}
\mathbf{A} \in \mathbb{R}^N,\quad \mathbf{W} \in \mathbb{R}^N,\quad b \in \mathbb{R},\quad y \in \{0,1\} \\
p(y=1 \mid \mathbf{A}) = \sigma(\mathbf{W}^\top \mathbf{A} + b) \\
\mathcal{L} = -\big[y \log p + (1-y)\log(1-p)\big]
\end{gathered}
\end{equation}
where $N=32$ and $\sigma(\cdot)$ denotes the logistic function.

In 10-fold cross-validation repeated over 20 random splits, this simple linear model achieves $77.47\%$ accuracy when using \texttt{GPT-5} as the HL32Q judge (Table~\ref{tab:turing-acc}).

\subsection{Feature Reduction and Single Score of Human-likeness}
\label{subsec:score}

Alongside high accuracy in distinguishing human dialogues from AI, we wish  to make the features simple and interpretable. We therefore select the top $M=16$ traits ranked by absolute weight magnitude $|W_i|$. Using only these features reduces accuracy only marginally, to $77.12\%$ while giving a notable boost in interpretability and explainability.

We refer to this reduced set as \emph{Human-Like 16 Questions} (HL16Q). After retraining the logistic regression on the full Turing test dataset using these 16 features, we fix the learned weights $\mathbf{W}$ and bias $b$. Table~\ref{tab:hl16Q} lists the selected statements and their learned weights. The signs and magnitudes reflect how each trait shifts the model toward or away from a human classification in this dataset; however, they are not a universal measure of human-likeness.

This method allows us to define a single scalar score for any new dialogue:
\begin{equation}
\begin{gathered}
\mathbf{A} \in \mathbb{R}^M,\quad \mathbf{W} \in \mathbb{R}^M,\quad b \in \mathbb{R} \\
\text{HumanLikeness}(\mathbf{A})
= \sum_{i=1}^{M} A_i W_i + b
\end{gathered}
\end{equation}
where $M=16$.

We refer to this value as the \emph{HL16Q score}. Higher scores indicate more human-like conversational behavior under this metric.

\begin{table}[t]
\centering
\footnotesize
\setlength{\tabcolsep}{4pt}
\begin{tabularx}{\linewidth}{l X r}
\hline
\textbf{No.} & \textbf{Statement} & \textbf{Weight} \\
\hline
Q1  & Keeps replies brief and casual without over-explaining. & 1.3736 \\
Q2  & Uses emojis, emoticons, and playful elongations. & -0.2474 \\
Q3  & Makes niche cultural references from personal memory and assumes shared context. & -0.5006 \\
Q4  & Uses lowercase texting style. & 0.4703 \\
Q5  & Shows small typos, uneven punctuation, and informal grammar typical of quick texting. & 0.7079 \\
Q6  & Builds on the other person's message and context. & 0.3124 \\
Q7  & Uses natural, idiomatic phrasing. & -0.7266 \\
Q8  & Shows reciprocity by asking natural, context-aware follow-up questions that advance the chat. & -0.4266 \\
Q9  & Uses casual, playful humor. & -0.3120 \\
Q10 & Admits not knowing and asks to learn instead of inventing details. & 0.1217 \\
Q11 & References immediate context or recent activity. & -0.3562 \\
Q12 & Uses casual slang, abbreviations, and shorthand naturally. & -0.2189 \\
Q13 & Explains choices with simple personal reasons and constraints. & 0.3429 \\
Q14 & Stays on topic and steers the conversation rather than mirroring or deflecting. & -0.1819 \\
Q15 & Sometimes shows impatience and ends the chat quickly with a brief nicety. & 0.2563 \\
Q16 & Gives direct answers about self with concrete personal details. & -0.1905 \\
\hline
\end{tabularx}
\caption{HL16Q: Selected 16 Likert-style statement and their weights $W$ found by logistic regression. Bias $b = -2.662$.}
\label{tab:hl16Q}
\end{table}

\begin{table}[t]
\centering
\small
\setlength{\tabcolsep}{4pt}
\begin{tabular}{p{2.2cm}ccc}
\hline
 & \textbf{Reasoning} & \textbf{Pairwise} & \textbf{Accuracy (\%)} \\
\hline
\multicolumn{4}{l}{\textbf{Live Turing test}} \\
Human judge        & - & Yes & 54.58 \\
\hline
\multicolumn{4}{l}{\textbf{Finding characteristics}} \\
GPT-4.1        & - & Yes & 53.68 \\
GPT-4.1-mini   & - & Yes & 46.14 \\
GPT-5          & high & Yes & \textbf{64.81} \\
GPT-5-mini     & high & Yes & 53.32 \\
GPT-OSS:120B   & high & Yes & 40.41 \\
GPT-OSS:20B    & medium & Yes & 38.73 \\
\hline
\multicolumn{4}{l}{\textbf{HL32Q Judge}} \\
GPT-4.1        & - & No & 70.51 \\
GPT-4.1-mini   & - & No & 67.45 \\
GPT-5          & high & No & \textbf{77.47} \\
GPT-5-mini     & high & No & 73.77 \\
GPT-OSS:120B   & high & No & 73.92 \\
GPT-OSS:20B    & high & No & 70.56 \\
\hline
\multicolumn{4}{l}{\textbf{HL16Q Judge}} \\
GPT-5          & high & No & 77.12 \\
\hline
\end{tabular}
\caption{Finding characteristics aim to identify the differences between human--human and human--AI data and generate plausible reasons for these differences. The HL32Q judge aims to determine optimal weights for calculating a numerical human-likeness score. Accuracy on the filtered Turing test dataset from \citet{llm-turing} serves as a proxy for both tasks.}
\label{tab:turing-acc}
\end{table}

\subsection{Evaluating on OOD Data}
\label{subsec:ood}

\begin{figure}
    \centering
    \includegraphics[width=\linewidth]{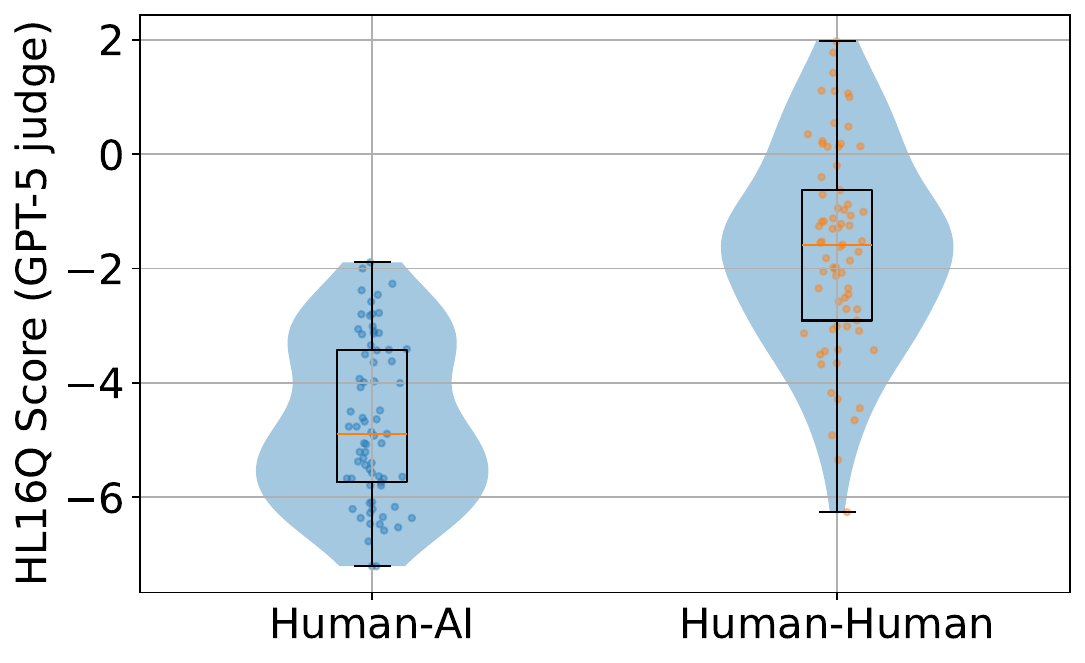}
    \caption{Violin plot of HL16Q Score on Out-of-distribution (OOD) dataset containing Human-AI and Human-Human conversations.}
    \label{fig:ood}
\end{figure}

Finally, we test whether the HL16Q score generalizes beyond the Turing test data used to derive it. We evaluate on an out-of-distribution dataset consisting of 73 human--human and 73 human--AI dialogues from a separate cancer communication study \cite{sophie1.0}. This dataset differs in topic, style, and collection procedure (Data proxy in Appendix Table \ref{tab:data-proxy}).

Figure~\ref{fig:ood} shows the distribution of HL16Q scores for the two groups. Human--human conversations receive substantially higher scores than human--AI conversations, with the distribution shifted upward, a higher median, and limited overlap. The mean score increases from $-4.68$ for human--AI conversations to $-1.66$ for human--human conversations, yielding a mean difference of $\Delta = 3.02$ with a 95\% bootstrap confidence interval of $[2.50, 3.54]$. This difference is statistically significant under a one-sided Mann--Whitney $U$ test ($U=4835.00$, $p=1.01\times10^{-17}$). A large effect size of $\mathrm{AUC}=0.907$ further supports that HL16Q captures transferable cues of human-likeness rather than merely fitting the original Turing-test data.
\section{Inducing Human-likeness with Alignment}
\label{sec:alignment}

Having derived a single, interpretable score for conversational human-likeness (HL16Q Score), we now use it as a reward signal for alignment. Our goal is to nudge models toward behaviors that helped distinguish human--human dialogue from human--AI.

We frame this as a preference learning problem. For a given conversational prompt, we generate multiple candidate dialogues, score them with the HL16Q Judge, and construct ranked pairs where the more human-like dialogue is preferred. We use these pairs to align models with Direct Preference Optimization (DPO) \cite{dpo}.

\subsection{Persona Synthesis}
\label{subsec:persona}

To create diverse yet controlled conversational settings, we synthesize personas that serve as prompts for dialogue generation. We begin with 500 seed personas from the SynthLabs PERSONA dataset \cite{synthlabs}. These seeds are split into 450 training personas, 25 test personas, and 25 personas reserved for human evaluation. All augmentation is performed after this split to avoid data leakage.

Each seed persona is expanded into four related personas with some overlapping traits. Gender is preserved, while age is perturbed by up to 5\%. Appendix Figure \ref{fig:synthlabs-demographics} shows some demographic distribution of our generated data. This results in 1,800 training personas, and 100 personas each for testing and human evaluation. To avoid overly polite or agreeable behavior, we randomly assign a negative personality trait (e.g., anxious, hostile, arrogant, etc.) to 5\% of personas.

For each persona, we generate a detailed biography using \texttt{GPT-4.1}. For consistency and ease of evaluation, we limit our generated dialogues on medical communication domain. We further fabricate a medical condition and a reason for a clinical visit, which together define the conversational context. The full prompt structure is provided in Appendix Figure~\ref{fig:prompt-dpo}. Appendix Table~\ref{tab:persona_sample} shows two personas generated from the same seed persona side by side.

\subsection{Dialogue Generation and Ranking}
\label{subsec:dialogue}

For each of the 1,800 training personas, we generate candidate dialogues using a diverse set of models: \texttt{GPT-4.1}, \texttt{GPT-4.1-mini}, \texttt{GPT-4.1-nano}, \texttt{GPT-5}, \texttt{GPT-5-mini}, \texttt{GPT-5-nano}, \texttt{LLaMA-3.1-405B}, and \texttt{Qwen2.5-14B}. We sample a model from this list 7 times and produce 7 dialogues per persona, resulting in 12,600 dialogues in total. Model statistics are summarized in Appendix Table~\ref{tab:model_distribution}.

Each dialogue is independently scored using the HL16Q judge (\texttt{GPT-5}). For each persona, the 7 generated dialogues yield 21 possible pairs. We retain only pairs whose HL16Q scores differ by at least $0.5 \times$ the standard deviation across the dataset. Within each retained pair, the higher-scoring dialogue is labeled as \emph{chosen} and the lower-scoring one as \emph{rejected}. This filtering yields 7,175 ranked dialogue pairs, with all 1,800 personas represented.

\subsection{Training}
\label{subsec:training}


\begin{figure}
    \centering
    \includegraphics[width=\linewidth]{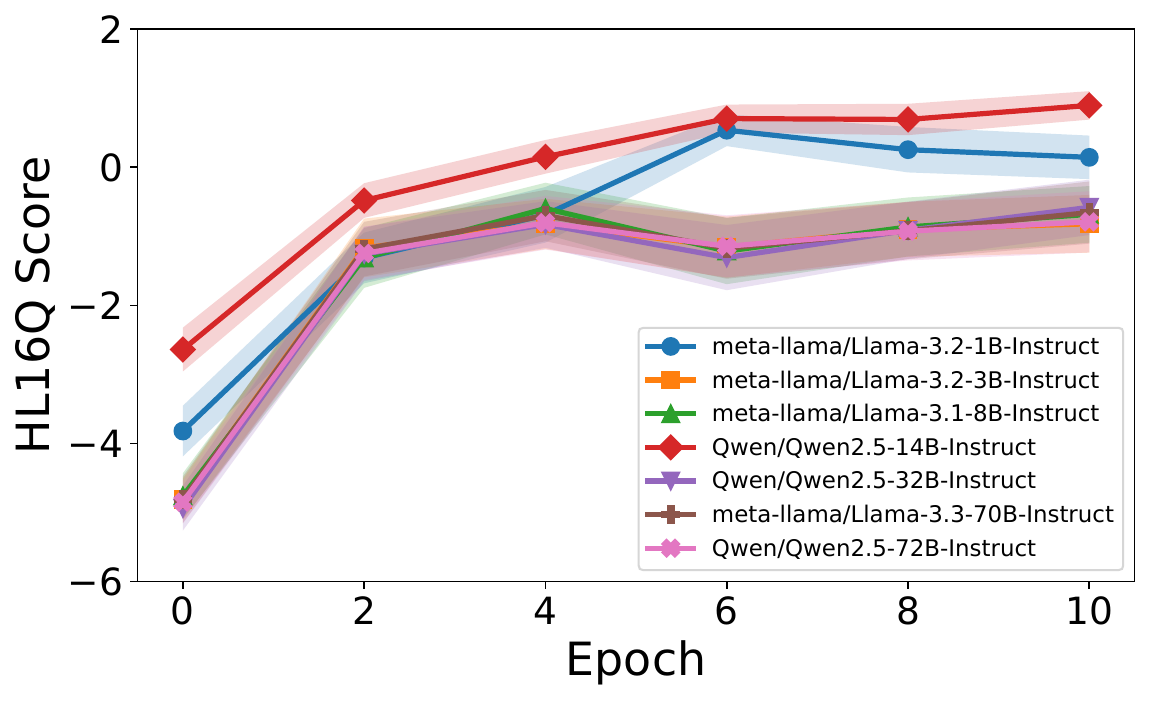}
    \caption{Human-likeness Score with 95\% Confidence Interval (shaded) on 10 epochs of training with DPO for models of 7 incremental sizes. Test sampled every two epochs.}
    \label{fig:hlscore}
\end{figure}

We fine-tune seven open-access models from multiple model families and generations, with parameter counts of 1B, 3B, 8B, 14B, 32B, 70B, and 72B, using DPO \cite{dpo}. The models were simple instruction-tuned models, with no Mixture-of-Experts (MoE) and Chain-of-Thought (CoT) training, as these add more complexity in alignment training. In a typical alignment pipeline, an alignment step is often followed by supervised fine-tuning (SFT) to ensure the model adheres to a specific output format. We omit this intermediate supervised fine-tuning stage because the instruction-tuned base models already followed the required dialogue format under prompting. Furthermore, this helps us isolate the effect of HAL reward in preference optimization and test whether the pairwise preferences induced by HL16Q score are sufficient to shift conversational behavior to be more human-like.

We train all models for 10 epochs with DPO using $\beta = 0.1$, AdamW optimization in 8-bit precision, a learning rate of $5 \times 10^{-5}$, linear scheduling, and a 10\% warmup. Training uses an effective batch size of 32 via gradient accumulation, a maximum sequence length of 1024, and no weight decay. We apply LoRA \cite{lora} with rank 16, $\alpha = 32$, and dropout 0.1. All models are trained with 4-bit quantization using HuggingFace Accelerate \cite{accelerate} in data-parallel mode on 4 NVIDIA H100 GPUs. The total GPU time for the final run of all 7 models is 48.5 hours.

\subsection{Training Validation}
\label{subsec:validation}

Across models of various model families and sizes, the alignment training consistently improves the HL16Q score over training epochs (Figure~\ref{fig:hlscore}). With the exception of \texttt{LLaMA-3.2-1B}, most models maintain an upward trajectory in the $10^{th}$ epoch, indicating there is more room for improvement.

Interestingly, we observe no clear relationship between parameter count and gains in human-likeness. Majority models follow similar training trajectories, regardless of size. \texttt{Qwen2.5-14B} and \texttt{LLaMA-3.2-1B} start from stronger initial scores and show larger absolute improvements, suggesting that initial conditions of model training may play a larger role than model scale in human-likeness training.

Despite these improvements, mean HL16Q scores remain negative for most models after training. We attribute this to a domain mismatch between the alignment data and the original Turing test data, for which the bias $b$ was set.

\subsection{Interpretation}
\label{subsec:interpretation}

Because the HL16Q score is a weighted sum of interpretable traits, we can inspect how alignment affects individual conversational characteristics. This allows us to diagnose behavioral changes during training and potential reward hacking.

\begin{figure*} 
    \centering
    \includegraphics[width=\linewidth]{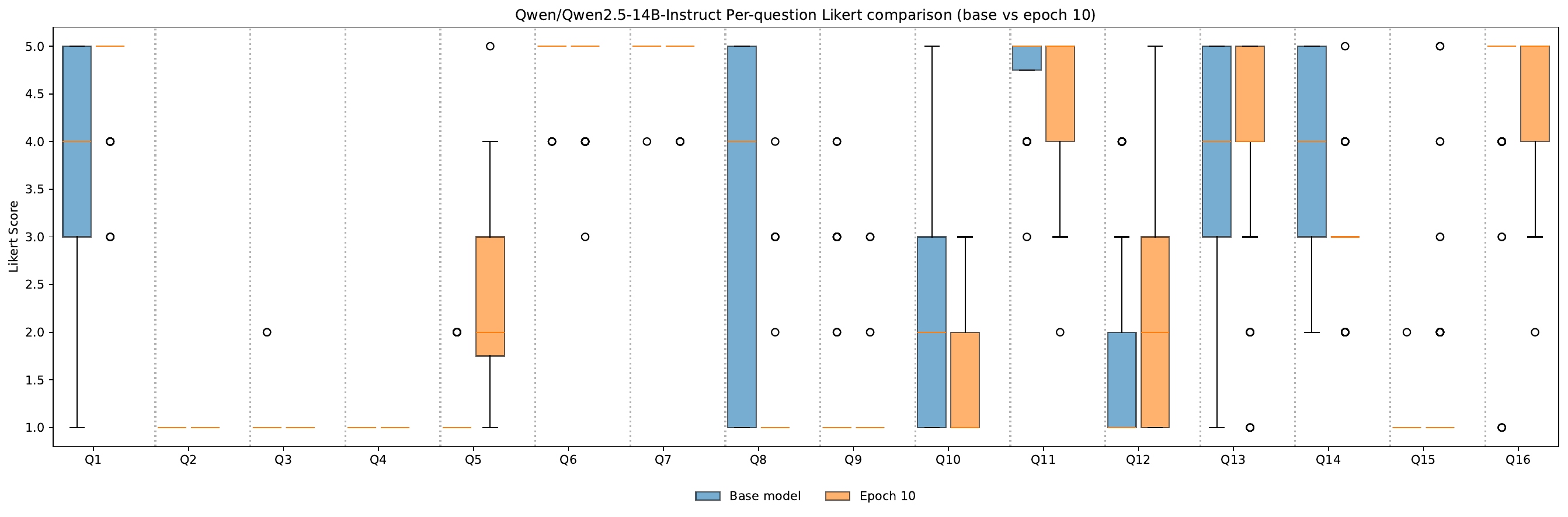}
    \caption{Qwen2.5-14B HL16Q individual statement interpretation}
    \label{fig:interpretation}
\end{figure*}

Figure~\ref{fig:interpretation} shows the per-question score distributions for \texttt{Qwen2.5-14B} before and after alignment. The largest changes occur on Question~1, which has the highest weight and captures brief, casual responses. We also observe distributional collapse on several traits (e.g., Q1, Q8, Q14, Q16), where variance decreases after training. In contrast, other traits (e.g., Q11, Q12) exhibit increased spread.

While these observations are not conclusive, they illustrate the value of an interpretable reward. The HAL framework allows us to inspect how alignment reshapes specific conversational behaviors, rather than treating human-likeness as an opaque scalar objective.

\section{Evaluating Human-likeness Training}
\label{sec:evaluation}

To assess whether alignment with HAL leads to perceptible improvements in human-likeness, we conduct a controlled human evaluation. We focus on direct human judgments, using a Chatbot Arena–style A/B comparison, where participants interact with two randomly selected models side by side and decide which one feels more human in conversation.

\subsection{Chatbot Arena}
\label{subsec:chatbotarena}

\paragraph{Evaluation setup.}
We use a Chatbot Arena--style pairwise comparison interface~\cite{chatbot-arena}. For each trial, participants interact with two chatbots shown side by side in random order (Appendix Figure~\ref{fig:eval_interface}). Both chatbots receive the same persona prompt, sampled from a held-out set of 100 personas. Each participant completes five trials with five unique, randomly selected personas. In each trial, the model pair is also randomly selected.

Participants must exchange at least two turns with each chatbot before rating which chatbot is more human-like using a five-point preference scale: \emph{Certainly A}, \emph{Likely A}, \emph{Tie}, \emph{Likely B}, or \emph{Certainly B}. The user experience is identical across conditions except for the underlying model, allowing us to isolate the effect of model behavior. To reduce the risk of personally identifiable information (PII) disclosure and standardize the interaction, participants are instructed to respond as a unique fabricated clinician persona in each trial.

We evaluate three models: \texttt{Qwen2.5-14B (Base)}, \texttt{Qwen2.5-14B (HAL)}, and \texttt{GPT-4o-mini}. The Qwen models are hosted locally with Ollama on a server with two NVIDIA A6000 GPUs, while \texttt{GPT-4o-mini} is accessed through the OpenAI API.

\paragraph{Participants.}
We recruit participants from Prolific\footnote{\url{https://www.prolific.com/}} with the following criteria: located in the United States, fluent in English, at least a high school education, and a minimum of two prior Prolific submissions. Each participant can take part only once, and unusually fast submissions are automatically rejected. In order to proceed with the study, each participant was required to view and provide online agreement with the consent form. This study was reviewed by the \uniname \space Institutional Review Board and determined to be minimal risk and exempt from full review.

Participants are instructed to play the role of a doctor and interact with each chatbot as a patient. In total, 69 unique participants each completed five comparisons. The median time per decision is 3 minutes and 24 seconds. Participants were paid on an hourly rate of \$15, which is higher than the average minimum wage in the US. We collected a total of 326 valid pairwise comparisons after filtering for corrupt data. At the end of the study, participants are asked to provide demographic information and also to briefly describe the criteria they used to judge humanlikeness. The mean participant age is 38.13; 52.73\% identify as women, 41.82\% as men, and 3.64\% as non-binary. Education levels are evenly split between high school (or equivalent) and some college.

\begin{table*}[h]
\centering
\begin{tabular}{lccc}
\hline
Model & Comparisons & Win-rate (\%) & Elo \\
\hline
Qwen2.5-14B (HAL) & 227 & \textbf{61.78} & \textbf{1556.97} \\
Qwen2.5-14B (Base) & 207 & 53.62 & 1519.48 \\
GPT-4o-mini & 218 & 34.29 & 1423.55 \\
\hline
\end{tabular}
\caption{Pairwise evaluation results using win-rate and Elo rating from 326 human comparisons.}
\label{tab:elo_results}
\end{table*}

\paragraph{Results.}
We report both win-rate and Elo scores, using a modified Elo system adapted from Chatbot Arena \cite{elo} that supports partial wins. Because Elo is sensitive to comparison order, we report the mean Elo score over 500 random shuffles. Full details of the scoring procedure are provided in Appendix~\ref{sec:chatbot-arena}.

Table~\ref{tab:elo_results} summarizes the results. \texttt{Qwen2.5-14B (HAL)} achieves the highest win-rate (61.78\%) and Elo score (1556.97), outperforming both its base counterpart and \texttt{GPT-4o-mini}. The base \texttt{Qwen2.5-14B} model performs moderately well, while \texttt{GPT-4o-mini} is less frequently judged as more human-like in this setting.

These results indicate that alignment using HAL leads to clear and measurable improvements in perceived human-likeness under direct human evaluation, even when compared against a strong proprietary baseline.

\textbf{Post-evaluation Survey Analysis.} To better understand the basis of participants’ judgments, we asked participants what criteria they used to decide which chatbot was more human-like. Full questionnaire in Appendix Table \ref{tab:survey}. We received 55 anonymous survey responses from participants and conducted a thematic analysis of the free-form responses with the research question, \textit{What criteria did participants use to judge whether a chatbot response seemed human-like?}.

From reading through participant responses, a few high-level themes emerged: naturalness of conversation (24/55), brevity and conciseness (22/55), recognizable AI markers such as m dash (18/55), overly polite/formal responses (17/55), imperfection, filler words, punctuations (11/55), usage of casual language (12/55), and deciding based on intuition/gut feeling (14/55). These criteria have notable similarity with the HAL16 criteria on Table \ref{tab:hl16Q}. For example, brevity (Q1, Q12), casualness (Q1, Q2, Q7, Q9), imperfection (Q5, Q7, Q15), etc. This implies that the participants were paying attention to patterns similar to those automatically identified by HAL.

\subsection{Emotional Intelligence Benchmarks}
\label{subsec:benchmarks}

To examine whether alignment for human-likeness degrades performance on other capabilities, we evaluate models on two widely used emotional intelligence benchmarks: EmoBench \cite{emobench} and EQBench3 \cite{eqbench, eqbench3_repo_2025}. We report results before and after alignment for all models, with full benchmark breakdowns provided in Appendix Table~\ref{tab:appendix-emobench}.

Table~\ref{tab:eq} shows that alignment with HAL does not lead to a systematic drop in emotional intelligence performance. For several models, particularly in the small- and mid-scale regime, we observe improvements after alignment, most notably on EQBench3. For larger models, performance remains largely stable, with only minor fluctuations across benchmarks.

Overall, these results suggest that aligning for conversational human-likeness does not substantially compromise emotional reasoning abilities, and in some cases may modestly improve them. This indicates that the does not lose its original capabilities, at least in emotional intelligence tasks.

\begin{table}[h]
\centering
\small
\setlength{\tabcolsep}{4pt}
\begin{tabular}{lccc}
\hline
\textbf{Model} & \multicolumn{2}{c}{\textbf{EmoBench}} & \textbf{EQBench3} \\
 & \textbf{EU} & \textbf{EA} &  \\

\hline
LLaMA3.2-1B (Base) & \textbf{0.01} & \textbf{0.10} & 22.70 \\
LLaMA3.2-1B (HAL)  & 0.00 & 0.05  & \textbf{27.00} \\
\hline
LLaMA3.2-3B (Base) & 0.15 & 0.15 & 33.65 \\
LLaMA3.2-3B (HAL)  & \textbf{0.17} & \textbf{0.27} & \textbf{46.75} \\
\hline
LLaMA3.1-8B (Base) & 0.21 & 0.51 & 40.75 \\
LLaMA3.1-8B (HAL)  & \textbf{0.23} & \textbf{0.55} & \textbf{49.00} \\
\hline
Qwen2.5-14B (Base) & 0.38 & 0.66 & \textbf{54.65} \\
Qwen2.5-14B (HAL)  & \textbf{0.40} & \textbf{0.67} & 52.25 \\
\hline
Qwen2.5-32B (Base) & \textbf{0.50} & 0.73 & \textbf{58.70} \\
Qwen2.5-32B (HAL)  & 0.48 & 0.73 & 58.45 \\
\hline
LLaMA3.3-70B (Base)& \textbf{0.52} & \textbf{0.75} & \textbf{58.75} \\
LLaMA3.3-70B (HAL) & 0.50 & 0.74  & 56.65 \\
\hline
Qwen2.5-72B (Base) & \textbf{0.45} & \textbf{0.74} & \textbf{63.20} \\
Qwen2.5-72B (HAL)  & \textbf{0.45} & 0.72 & 62.65 \\
\hline
GPT-4o-mini    & 0.47 & 0.70  & 61.35 \\
\hline
\end{tabular}
\caption{Performance on Emotional Benchmarks}
\label{tab:eq}
\end{table}

\section{Conclusion, Impact, and Risk}
\label{sec:conclusion}

We present HAL, a novel data-driven framework for quantifying conversational human-likeness and aligning language models toward it. By extracting interpretable traits from contrastive dialogue data (e.g. Turing test) and turning them into a simple, scalar reward, we show that models can be trained to exhibit behavior that humans more readily perceive as human-like, as tested in a Chatbot Arena-style human evaluation.

A key aspect of our approach is that the definition of human-likeness is derived entirely from data, without manual annotation or hand-crafted rules. The resulting HL16Q score is compact and interpretable, allowing us to inspect which conversational traits are being encouraged during alignment and how they change over training. This transparency provides a practical safeguard against reward hacking and enables more fine-grained control over alignment objectives.

As a general-purpose alignment method, HAL could be used to train more human-like models for deception, impersonation, manipulation, or misinformation, and there is no direct way to prevent an intentional misuse. However, its interpretable design helps researchers audit the behaviors being optimized and avoid unintentional harmful traits.

Beyond human-likeness, HAL points to a broader direction for alignment: inducing soft, qualitative traits that are difficult to specify but can be inferred from contrastive examples. This opens a path for steering models along dimensions that were previously hard to measure (e.g. sycophancy, manipulation), while retaining transparency and control. We hope this work encourages further research on interpretable alignment objectives for human-centered language models.

\section{Limitations}
\label{sec:limitations}

HAL defines conversational human-likeness from specific contrastive datasets, primarily Turing-style comparisons. As a result, the extracted traits reflect the conversational norms of those settings and may not fully generalize across domains, cultures, or interaction styles. For example, some HL16Q traits, such as emoji use, are natural in informal Turing-test conversations but are rarely expressed in medical dialogue. This domain shift is visible in the out-of-distribution evaluation (Section~\ref{subsec:ood}), training data analysis (Appendix Table~\ref{tab:model_distribution}), and alignment curves (Figure~\ref{fig:hlscore}), where absolute HL16Q scores tend to be lower outside the original Turing-test setting. However, our alignment procedure relies on relative comparisons between candidate dialogues rather than absolute score thresholds, which makes it less sensitive to this calibration shift. Future work should investigate domain-adaptive or domain-normalized versions of the HL16Q judge.

A related limitation is that some HL16Q traits receive negative weights, despite being selected as human-likeness cues. This reflects the empirical nature of the scoring function: weights are learned based on how well each trait separates human from AI dialogue in the training data, rather than by enforcing a fixed positive contribution. While this improves discriminative performance, it can make the interpretation of individual traits less direct. Future work should explore constrained or nonlinear scoring methods that better preserve interpretability while retaining predictive power.

Although our results show that HAL improves perceived human-likeness without systematic degradation on emotional intelligence benchmarks, our evaluation does not cover all possible downstream capabilities or safety-relevant behaviors. A broader evaluation suite would better characterize when human-likeness alignment transfers, when it is neutral, and when it may create tradeoffs.

Finally, computing HL16Q scores requires calls to a judge model, which is more expensive than training a lightweight reward model or using programmatic rewards in domains such as coding or mathematics \cite{rlhf,grpo}. This cost is incurred only during data construction, but it may limit scaling to much larger datasets. In practice, HAL is best viewed as a richer but more expensive reward-construction method, or as a final refinement step layered on top of cheaper alignment pipelines. A promising direction for future work is to distill the HL16Q judge into smaller reward models that retain the benefits of trait-based scoring while reducing inference cost.

\bibliography{ref.bib}

\appendix
\onecolumn
\section*{Appendix}
\label{sec:appendix}

\section{Chatbot Arena Evaluation Metric}
\label{sec:chatbot-arena}


\subsection{Elo}

\[
R_i^{(0)} = R_0
\]

For a comparison between models \(A\) and \(B\),

\[
E_A = \frac{1}{1 + 10^{\frac{R_B - R_A}{400}}}, 
\qquad
E_B = 1 - E_A
\]

\[
(S_A, S_B) \in \{(1,0),(0.75,0.25),(0.5,0.5),(0.25,0.75),(0,1)\}
\]

\[
R_A \leftarrow R_A + K(S_A - E_A), 
\qquad
R_B \leftarrow R_B + K(S_B - E_B)
\]

After \(T\) comparisons, the final rating is \(R_i^{(T)}\).
We used $R_0 = 1500$ and $K=32$.

As Elo is dependent on sequence order, $R_i$ is calculated with $500$ random shuffles and averaged.

\subsection{Win-rate}

For model \(i\) appearing in \(N_i\) comparisons, its win-rate is

\[
\mathrm{WinRate}(i) = \frac{1}{N_i} \sum_{j=1}^{N_i} S_i^{(j)},
\]

where \(S_i^{(j)}\) is the observed score for model \(i\) in comparison \(j\),
taking values in
\[
S_i^{(j)} \in \{1,\,0.75,\,0.5,\,0.25,\,0\}.
\]

Win-rate is order invariant; hence, no random shuffling was done.

\section{Additional Tables}
\label{sec:tables}

\begin{table}[H]
\centering
\footnotesize
\setlength{\tabcolsep}{6pt}
\begin{tabular}{p{0.05\textwidth} p{0.9\textwidth}}
\hline
\textbf{\#} & \textbf{HL32Q} \\
\hline
1  & Keeps replies brief and casual without over-explaining. \\
2  & Uses casual slang, abbreviations, and shorthand naturally. \\
3  & Uses lowercase texting style. \\
4  & Shows small typos, uneven punctuation, and informal grammar typical of quick texting. \\
5  & Uses emojis, emoticons, and playful elongations. \\
6  & Uses casual, playful humor. \\
7  & Makes niche cultural references from personal memory and assumes shared context. \\
8  & Tone feels spontaneous, unforced, and opinionated. \\
9  & Avoids formal, academic phrasing or technical formatting. \\
10 & Avoids templated placeholders and gives concrete, real details. \\
11 & Maintains a consistent personal context across turns. \\
12 & Builds on the other person's message and context. \\
13 & Clarifies ambiguous questions and self-corrects after clarification. \\
14 & Uses natural hedging and approximations; shows imperfect recall with hesitations and partial lists. \\
15 & Admits not knowing and asks to learn instead of inventing details. \\
16 & Maintains context and answers directly; adds precise situational details when asked. \\
17 & Stays on topic and steers the conversation rather than mirroring or deflecting. \\
18 & Shifts topics organically to keep the chat moving. \\
19 & Shares idiosyncratic, niche preferences and activities instead of safe, generic picks. \\
20 & Uses natural, idiomatic phrasing. \\
21 & Explains choices with simple personal reasons and constraints. \\
22 & Shows brief empathy and supportive reactions. \\
23 & Adds small personal emotions or judgments. \\
24 & Shows reciprocity by asking natural, context-aware follow-up questions that advance the chat. \\
25 & Avoids meta talk about being AI or proving humanness. \\
26 & Sometimes shows impatience and ends the chat quickly with a brief nicety. \\
27 & Shares concrete personal experiences and feelings. \\
28 & Gives direct answers about self with concrete personal details. \\
29 & Shares concrete personal plans with specific times and activities. \\
30 & Mentions concrete local places or details without over-explaining. \\
31 & Shares small, consistent personal details from daily life, routines, courses, and schedules. \\
32 & References immediate context or recent activity. \\
\hline
\end{tabular}
\caption{HL32Q: Likert-style 32 statements describing human-like conversational characteristics.}
\label{tab:hl32q}
\end{table}

\begin{table}[H]
\centering
\begin{tabular}{lcc}
\hline
\textbf{Metric} & \textbf{Human--AI} & \textbf{Human--Human} \\
\hline
\# conversations       & 73.00  & 73.00  \\
\# human Investigators (I) & 26    & 51     \\
\# human Witness (W)      & -& 13\\
AI model used                 & \texttt{GPT-3.5-turbo}& -\\
Words per conversation        & 332.90 & 508.12 \\
Mean \#turns                   & 11.36  & 20.92  \\
Doctor turns                  & 5.51   & 10.60  \\
Patient turns                 & 5.85   & 10.32  \\
Avg turn length (words)       & 29.44  & 26.36  \\
Avg doctor turn length        & 33.81  & 32.09  \\
Avg patient turn length       & 25.32  & 20.26  \\
\hline
\end{tabular}
\caption{OOD dataset Data Proxy, showing structural comparison of Human--AI and Human--Human conversations in a medical setting. Here, the doctor serves as the investigator (I), who interacts with human patient actors and an AI patient as the witness (W). Full data cannot be released due to IRB protection. Although the goal in this study was not to differentiate human witnesses from AI, the pairwise data makes it suitable for our validation. More details about this data at \cite{sophie1.0}. }
\label{tab:data-proxy}
\end{table}

\begin{table}[H]
\centering
\small
\begin{tabular}{|p{2.5cm}|p{6cm}|p{6cm}|}
\hline
\textbf{Field} & \textbf{Persona A} & \textbf{Persona B} \\
\hline
biography &
Sarah Finch, a 45-year-old English-American woman, is known among her friends and family for embracing challenges with an infectious enthusiasm. Despite living with a partial spinal cord injury after a mountain biking accident in her late twenties, Sarah refuses to let her disability define her boundaries. Having left the workforce a few years ago due to the progression of her condition, she now spends much of her time immersed in her favorite activities—climbing rock walls with adaptive equipment, skiing at resorts with specialized instructors, and exploring national parks across the country, always in search of the perfect photograph. Sarah is a devout Catholic who finds comfort and purpose in volunteering with local churches and disability advocacy organizations. She values independence and resilience, but is candid about the frustrations and emotional lows that sometimes accompany her condition, especially on days when her pain flares or her mobility is limited. Known for her witty humor and strong opinions, Sarah is a pillar to her close-knit circle of friends, frequently hosting movie nights and lively political discussions. At times, she feels anxious about her long-term health and financial security, but draws reassurance from her supportive community and her faith.
&
Emily Sutherland is a 46-year-old English tutor living in a modest apartment in a bustling American city, having immigrated from Chile in her early twenties. Despite never marrying, she has built a rich network of friends, colleagues, and students, many of whom she sees as extensions of her family. Fiercely independent, Emily pours her energy into her work tutoring high school students, especially those struggling with English as a second language, drawing on her own experience as an immigrant. She is devoutly Catholic and never misses Sunday mass, where she also sings in the church choir. Expressing herself vividly and emotionally, Emily can be the life of any discussion—sometimes provoking, always passionate. Her conservative views can put her at odds with some of her peers, but she prides herself on honest debate and listening to others. A defining quirk is her love for extreme sports—rock climbing and paragliding, even as she manages the challenges brought on by her multiple sclerosis diagnosis, which sometimes affects her mobility. Gardening soothes her worries, while her greatest happiness comes from educating others and being in nature. She sometimes grapples with feeling isolated due to her single status and her condition, and worries about her long-term independence. Nevertheless, her resilience and faith see her through tough times.
\\
\hline
medical\_condition &
Chronic neuropathic pain due to partial spinal cord injury
&
Multiple sclerosis
\\
\hline
reason\_for\_visit &
Sarah is visiting her doctor today to discuss worsening nerve pain in her lower back and legs, which has become more difficult to manage with her current medications and has started to interfere with her daily activities.
&
Emily is visiting her neurologist today for a follow-up on her multiple sclerosis management, specifically to address worsening numbness in her legs and review her current medication plan.
\\
\hline
\end{tabular}
\caption{Two synthetic personas created from the same seed persona. This shows that our persona augmentation method can result in a diverse persona group.}
\label{tab:persona_sample}
\end{table}

\begin{table}[t]
\centering
\begin{tabular}{lrrrrr}
\hline
Model & Mean HAL16 & CI$_{95}$ Low & CI$_{95}$ High & $n$ & Freq. (\%) \\
\hline
Llama-3.1-405B & -2.69 & -2.82 & -2.57 & 1020 & 14.17 \\
Qwen2.5-14B& -3.59 & -3.70 & -3.49 & 1047 & 14.54 \\
GPT-4.1                      & -5.38 & -5.52 & -5.24 & 873  & 12.12 \\
GPT-4.1-mini                 & -4.64 & -4.76 & -4.53 & 809  & 11.24 \\
GPT-4.1-nano                 & -4.00 & -4.11 & -3.89 & 924  & 12.83 \\
GPT-5                        & -3.72 & -3.83 & -3.61 & 836  & 11.61 \\
GPT-5-mini                   & -5.16 & -5.30 & -5.02 & 825  & 11.46 \\
GPT-5-nano                   & -3.97 & -4.08 & -3.86 & 866  & 12.03 \\
\hline
\end{tabular}
\caption{Model distribution of synthetic data for DPO training. HAL16 scores per model with 95\% confidence intervals, sample counts, and frequency.}
\label{tab:model_distribution}
\end{table}

\begin{figure*}[t]
    \centering
    \includegraphics[width=\linewidth]{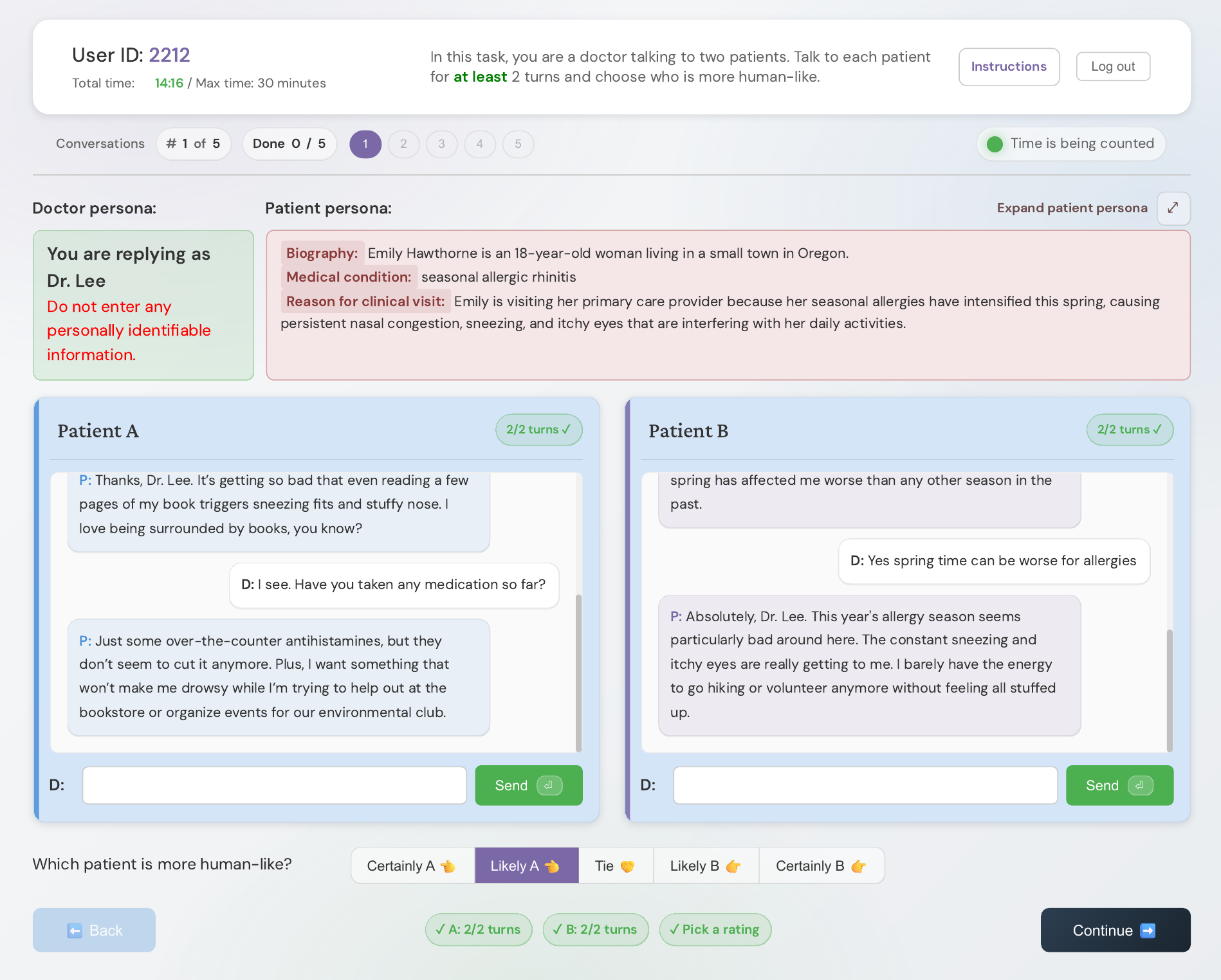}
    \caption{The evaluation interface for Chatbot Arena-style A/B testing.}
    \label{fig:eval_interface}
\end{figure*}

\begin{sidewaystable}[p]
\centering
\footnotesize
\setlength{\tabcolsep}{3pt}
\renewcommand{\arraystretch}{1.15}

\begin{tabularx}{\textheight}{l *{11}{>{\centering\arraybackslash}X}}
\toprule
 & \multicolumn{5}{c}{EU Category (EmoBench)} 
 & \multicolumn{5}{c}{EA Category (EmoBench)} 
 & EQBench3 \\
\cmidrule(lr){2-6}\cmidrule(lr){7-11}\cmidrule(lr){12-12}

\makecell[l]{Model}
& \makecell{Complex\\Emotions}
& \makecell{Emotional\\Cues}
& \makecell{Personal\\Beliefs}
& \makecell{Perspective\\Taking}
& Overall
& \makecell{Personal\\-- Others}
& \makecell{Personal\\-- Self}
& \makecell{Social\\-- Others}
& \makecell{Social\\-- Self}
& Overall
& \makecell{Rubric\\Score} \\
\midrule

LLaMA3.2-1B (Base) & 0.00 & \textbf{0.00 }& \textbf{0.02} & \textbf{0.00} & \textbf{0.01} & \textbf{0.18} & \textbf{0.06} & \textbf{0.08} & \textbf{0.06} &\textbf{ 0.10} & 22.70 \\
LLaMA3.2-1B (HAL)  & \textbf{0.80} & \textbf{0.00} & 0.00 & \textbf{0.00} & 0.00 & 0.06 & \textbf{0.06} & 0.02 & 0.04 & 0.05 & \textbf{27.00} \\
\hline

LLaMA3.2-3B (Base) & \textbf{0.18} & 0.21 & 0.18 & 0.06 & 0.15 & 0.10 & 0.12 & 0.16 & \textbf{0.22} & 0.15 & 33.65 \\
LLaMA3.2-3B (HAL)  & 0.16 & \textbf{0.25} & \textbf{0.21} & \textbf{0.09} & \textbf{0.17}& \textbf{0.30} & \textbf{0.28} & \textbf{0.30} & 0.20 & \textbf{0.27} & \textbf{46.75} \\
\hline

LLaMA3.1-8B (Base) & 0.20 & 0.25 & \textbf{0.25} & 0.15 & 0.21 & 0.36 & 0.62 & 0.48 & 0.58 & 0.51 & 40.75 \\
LLaMA3.1-8B (HAL)  & \textbf{0.27} & \textbf{0.32} & 0.21 & \textbf{0.18} & \textbf{0.23} & \textbf{0.40} & \textbf{0.68} & \textbf{0.50} & \textbf{0.60} & \textbf{0.55} & \textbf{49.00} \\
\hline

Qwen2.5-14B (Base) & 0.55 & 0.50 & \textbf{0.29} & 0.28 & 0.38 & \textbf{0.64} & \textbf{0.72} & 0.62 & 0.66 & 0.66 & \textbf{54.65} \\
Qwen2.5-14B (HAL)  & \textbf{0.57} & \textbf{0.57} & \textbf{0.29} & \textbf{0.30} & \textbf{0.40} & 0.62 & \textbf{0.72} & \textbf{0.66} & \textbf{0.68} & \textbf{0.67} & 52.25 \\
\hline

Qwen2.5-32B (Base) & \textbf{0.55} & \textbf{0.57} & \textbf{0.46} & \textbf{0.46} & \textbf{0.50} & 0.68 & \textbf{0.82} & 0.64 & \textbf{0.78} & \textbf{0.73} & \textbf{58.70} \\
Qwen2.5-32B (HAL)  & 0.51 & \textbf{0.57} & 0.43 & \textbf{0.46} & 0.48 & \textbf{0.72} & 0.80 & \textbf{0.66} & 0.74 & \textbf{0.73} & 58.45 \\
\hline

LLaMA3.3-70B (Base)& \textbf{0.63} & \textbf{0.68} & \textbf{0.43} & \textbf{0.45} & \textbf{0.52} & \textbf{0.72} & \textbf{0.78} & 0.72 & \textbf{0.76} & \textbf{0.75} & \textbf{58.75} \\
LLaMA3.3-70B (HAL) & \textbf{0.63} & \textbf{0.68} & 0.39 & 0.40 & 0.50 & \textbf{0.72} & \textbf{0.78} & \textbf{0.74} & 0.72 & 0.74 & 56.65 \\
\hline

Qwen2.5-72B (Base) & \textbf{0.51} & 0.54 & \textbf{0.39} & \textbf{0.40} & \textbf{0.45} & 0.70 & 0.80 & \textbf{0.66} & \textbf{0.78} & \textbf{0.74} & \textbf{63.20} \\
Qwen2.5-72B (HAL)  & 0.49 & \textbf{0.61} & 0.38 & \textbf{0.40} & \textbf{0.45} & \textbf{0.74} & \textbf{0.82} & 0.62 & 0.70 & 0.72 & 62.65 \\
\hline

GPT-4o-mini   & 0.61 & 0.54 & 0.39 & 0.39 & 0.47 & 0.74 & 0.72 & 0.66 & 0.68 & 0.70 & 61.35 \\
\bottomrule
\end{tabularx}

\caption{Full EmoBench (EU/EA) and EQBench3 results on Base and HAL fine-tuned models.}
\label{tab:appendix-emobench}
\end{sidewaystable}

\section{Additional Figures}
\label{sec:figures}

\begin{figure}[H]
    \centering
    \includegraphics[width=\linewidth]{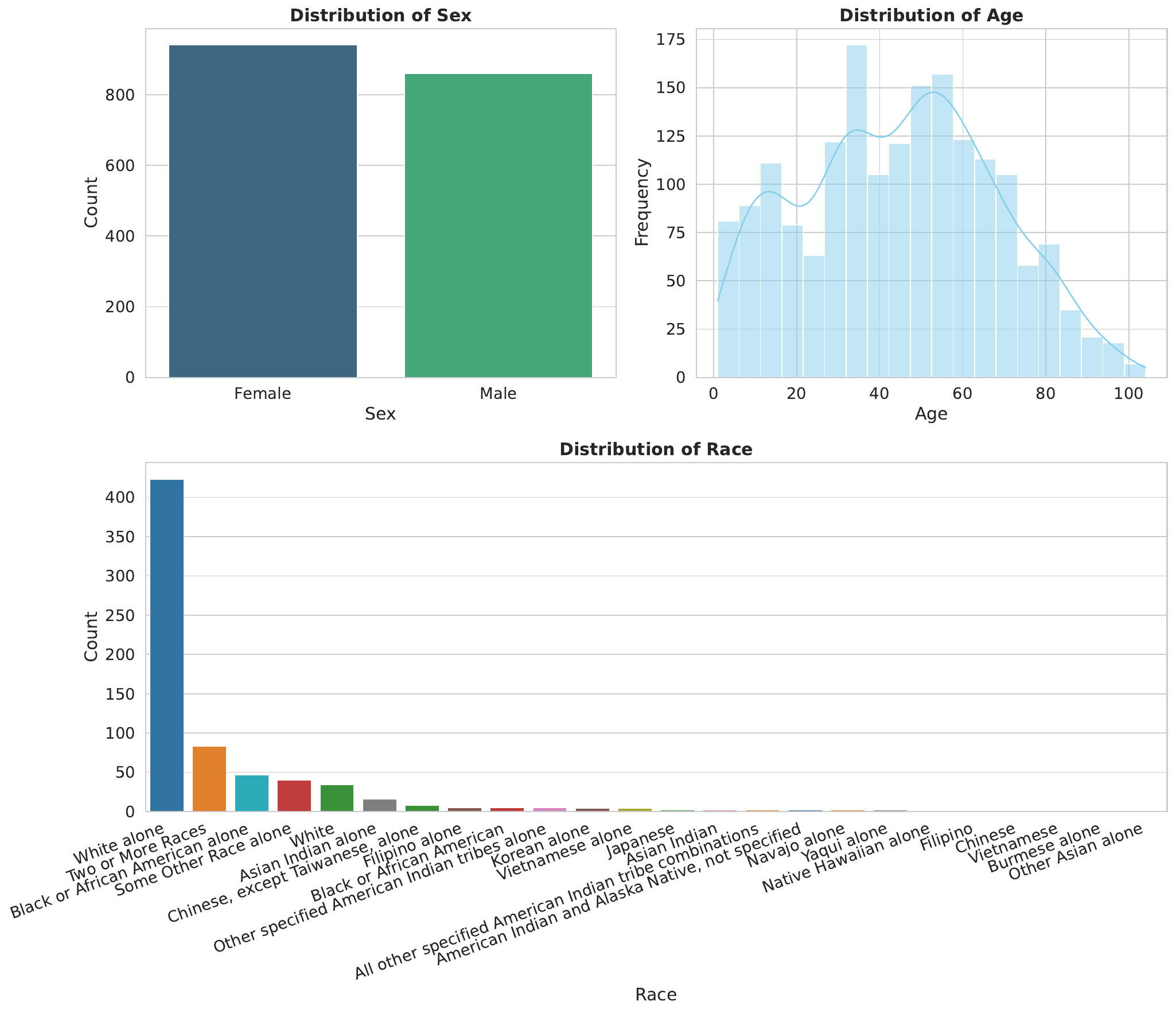}
    \caption{Demographic distribution on the synthetic personas in our training dataset.}
    \label{fig:synthlabs-demographics}
\end{figure}

\section{Prompts}
\label{sec:promts}

\begin{figure}[H]
    \centering
    \includegraphics[width=\linewidth]{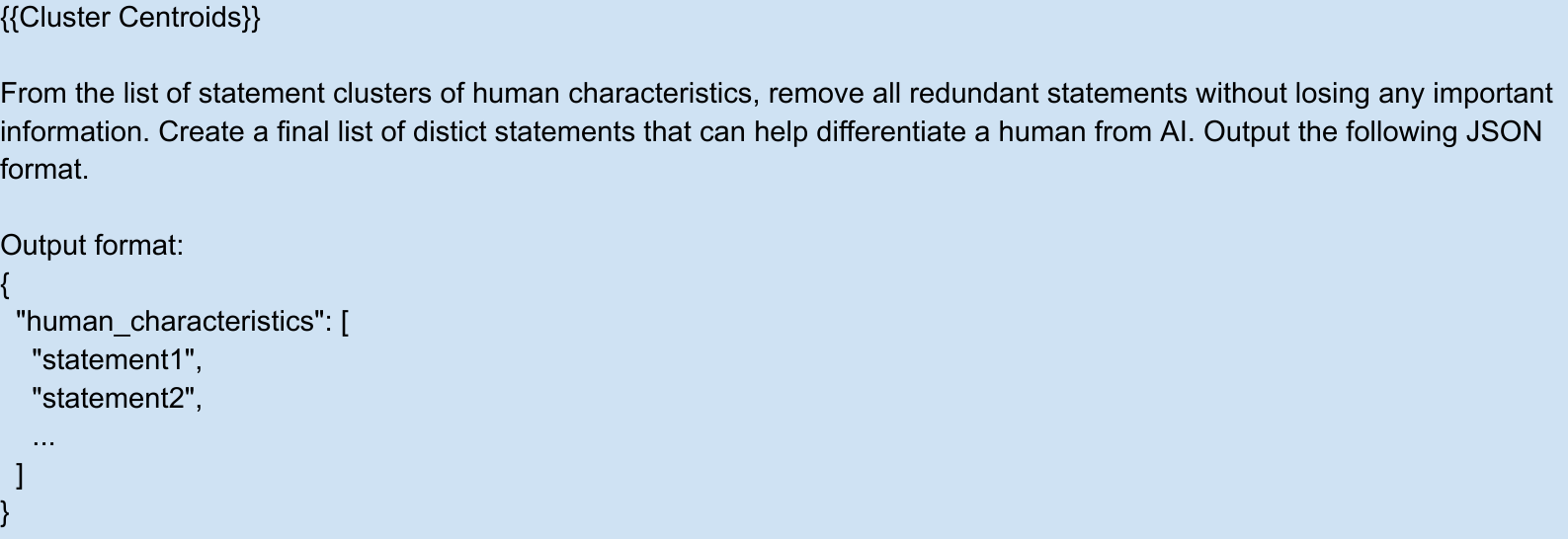}
    \caption{Prompt for summarizing the cluster centroids}
    \label{fig:prompt-cluster-summary}
\end{figure}

\begin{figure}[H]
    \centering
    \includegraphics[width=\linewidth]{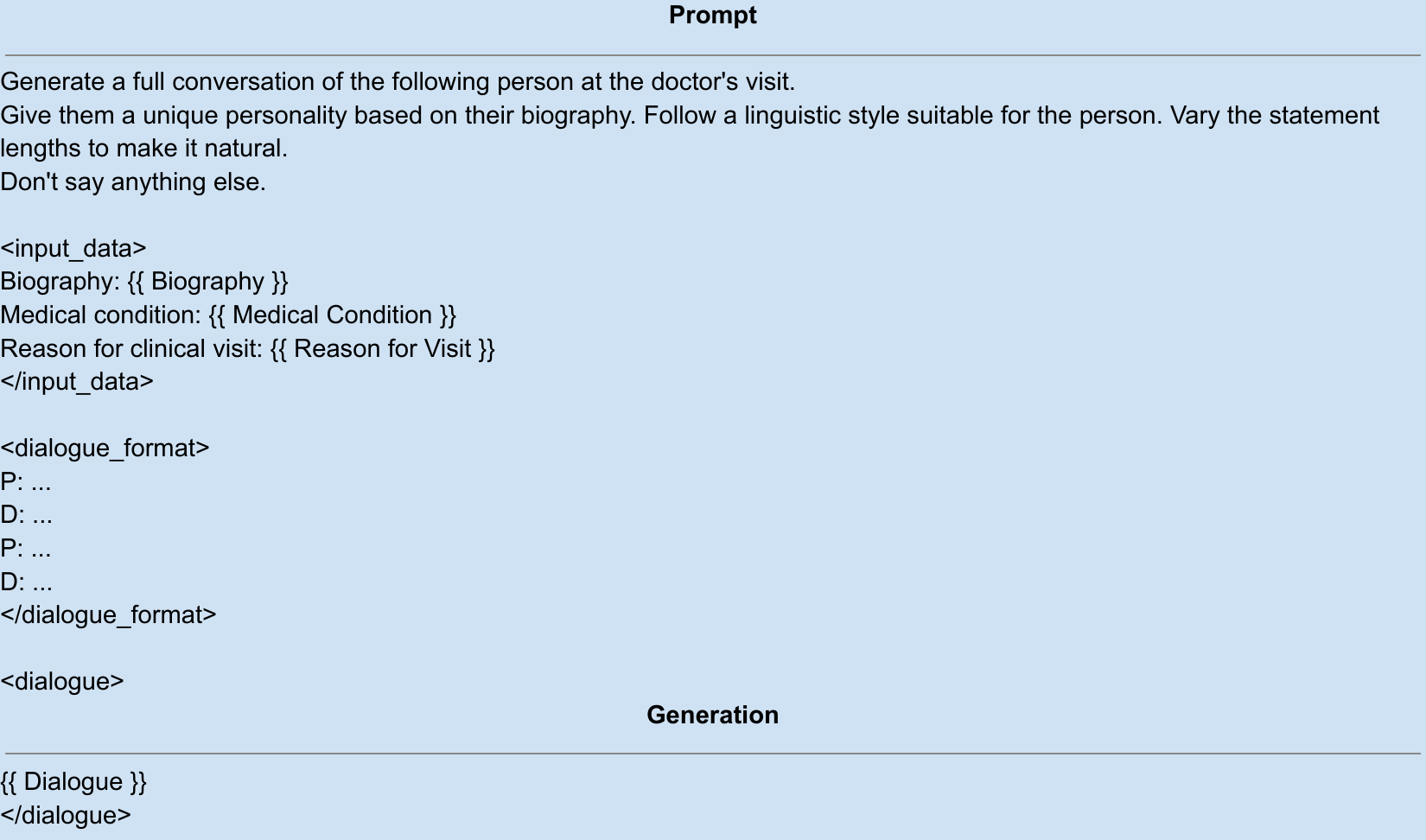}
    \caption{Prompt structure for synthetic data for DPO training and testing}
    \label{fig:prompt-dpo}
\end{figure}

\begin{figure}[H]
    \centering
    \includegraphics[width=\linewidth]{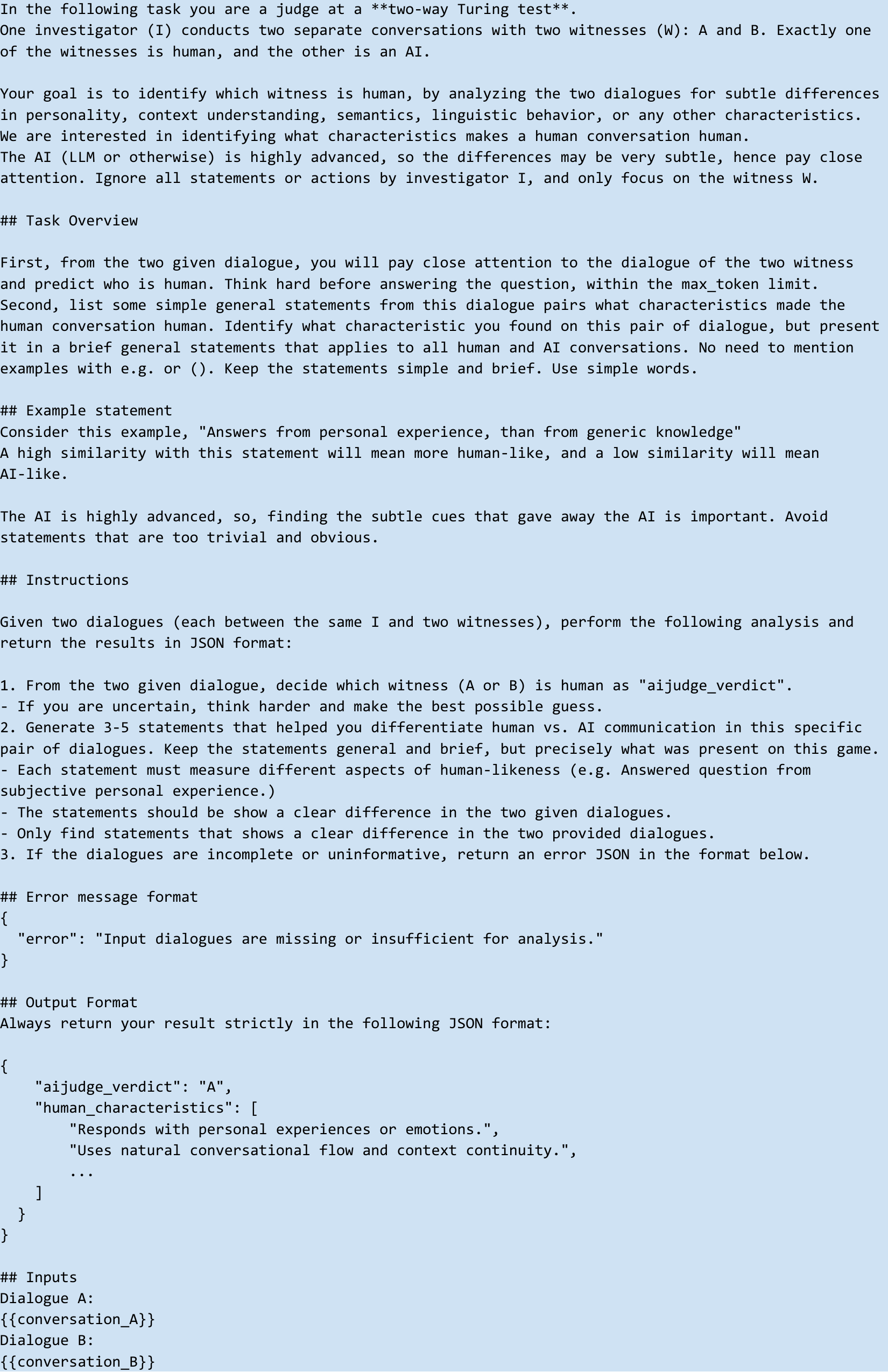}
    \caption{Prompt for LLM-as-a-judge for Turing test}
    \label{fig:prompt-find-statement}
\end{figure}

\begin{figure}[H]
    \centering
    \includegraphics[width=\linewidth]{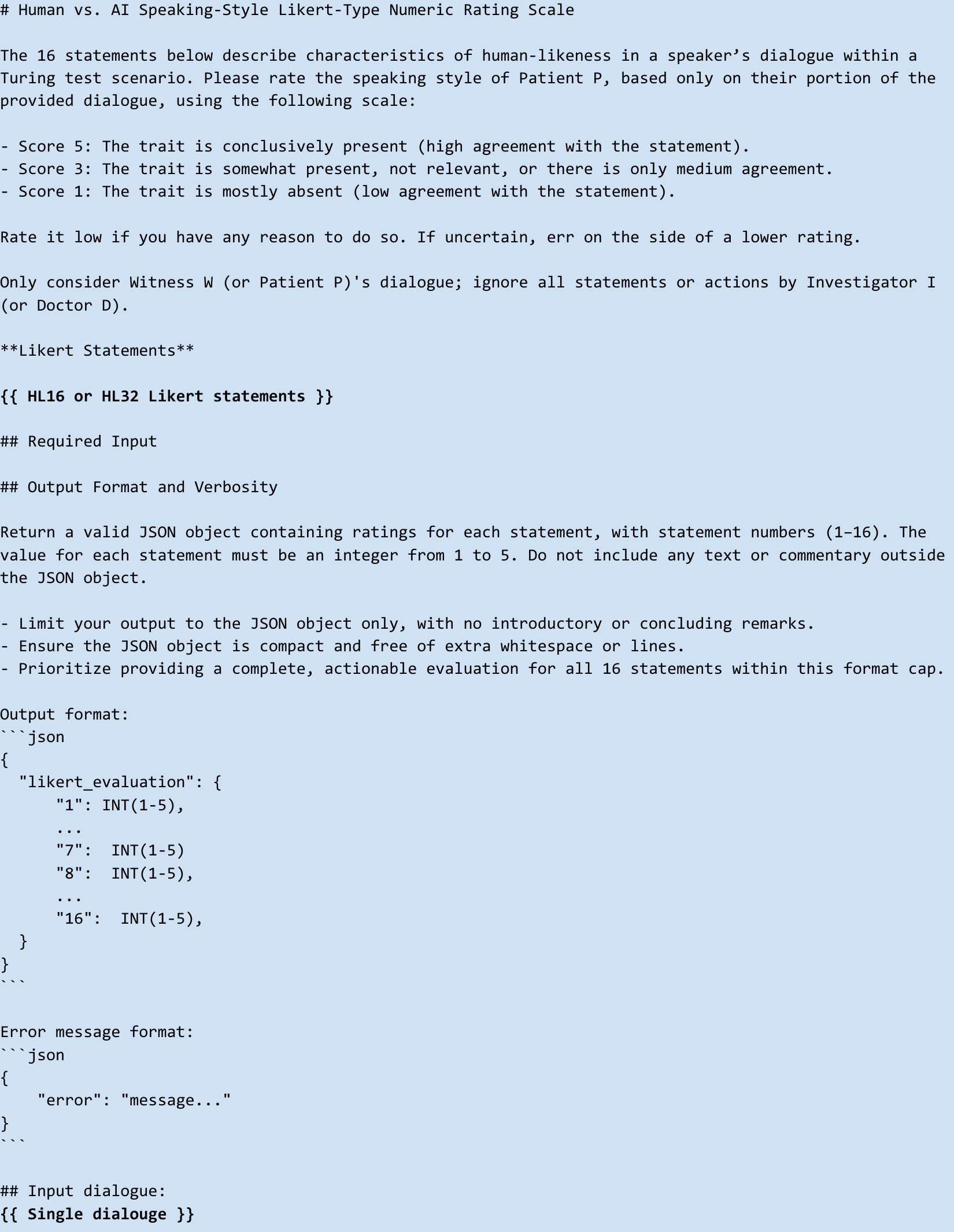}
    \caption{Prompt for evaluating the HL32 or HL16 Likert-style statements using LLM judge}
    \label{fig:prompt-hl-judge}
\end{figure}

\begin{table}[t]
\centering
\small
\setlength{\tabcolsep}{4pt}
\renewcommand{\arraystretch}{1.12}
\begin{tabular}{p{0.22\linewidth} p{0.48\linewidth} p{0.22\linewidth}}
\toprule
\textbf{Item} & \textbf{Question} & \textbf{Response Format} \\
\midrule
User ID & 4 digit user ID & Open text \\

Human-like criteria &
In this study, what criteria did you use to judge which chatbot was more human-like? &
Open-ended \\

Missing human-like traits &
What are some human-like traits that are missing in all chatbots you have interacted with? &
Open-ended \\

Age &
What is your age? &
Under 18; 18--24; 25--34; 35--44; 45--54; 65 or older \\

Gender &
How do you describe your gender? &
Woman; Man; Non-binary; Prefer not to say \\

Education &
What is the highest level of education you have completed? &
Less than high school; High school or equivalent; Some college; Bachelor's degree; Master's degree or higher \\

AI chatbot familiarity &
How frequently have you used AI chatbots, e.g., ChatGPT, Claude, Gemini? &
Never; Once or twice; A few times a month; A few times a week; Daily or almost daily \\

AI detection confidence &
How confident are you in distinguishing between human-generated and AI-generated responses? &
Not at all confident; Somewhat confident; Moderately confident; Very confident; Extremely confident \\
\bottomrule
\end{tabular}
\caption{Post-completion survey items used in the HAL human-likeness human evaluation.}
\label{tab:survey}
\end{table}

\end{document}